\documentclass[conference]{IEEEtran}
\IEEEoverridecommandlockouts
\usepackage{amsmath,amssymb,amsfonts}
\usepackage{algorithmic}
\usepackage{graphicx}
\usepackage{textcomp}
\usepackage{xcolor}
\def\BibTeX{{\rm B\kern-.05em{\sc i\kern-.025em b}\kern-.08em
    T\kern-.1667em\lower.7ex\hbox{E}\kern-.125emX}}

\usepackage{url}

\usepackage{cite}  

\usepackage{float}
\usepackage{balance}

\begin{document}

\title{Question-Answering System for Bangla: Fine-tuning BERT-Bangla for a Closed Domain\\
}
\author{\IEEEauthorblockN{Subal Chandra Roy}
\IEEEauthorblockA{\textit{Department of Computer Science and Engineering} \\
\textit{Khulna University of Engineering \& Technology}\\
Khulna, Bangladesh \\
subalroy5561@gmail.com}
\and
\IEEEauthorblockN{Md. Motaleb Hossen Manik}
\IEEEauthorblockA{\textit{Department of Computer Science and Engineering} \\
\textit{Khulna University of Engineering \& Technology}\\
Khulna, Bangladesh \\
mh.manik@cse.kuet.ac.bd}
}

\maketitle

\begin{abstract}
Question-answering systems for Bengali have seen limited development, particularly in domain-specific applications. Leveraging advancements in natural language processing, this paper explores a fine-tuned BERT-Bangla model to address this gap. It presents the development of a question-answering system for Bengali using a fine-tuned BERT-Bangla model in a closed domain. The dataset was sourced from Khulna University of Engineering \& Technology's (KUET) website and other relevant texts. The system was trained and evaluated with 2500 question-answer pairs generated from curated data. Key metrics, including the Exact Match (EM) score and F1 score, were used for evaluation, achieving scores of 55.26\% and 74.21\%, respectively. The results demonstrate promising potential for domain-specific Bengali question-answering systems. Further refinements are needed to improve performance for more complex queries.
\end{abstract}


\section{Introduction}
The rapid advancement of artificial intelligence (AI) in recent years has significantly improved the ability of machines to understand and process human languages \cite{ali2021ai}. As a result, question-answering (QA) systems have emerged as critical tools in natural language processing (NLP) \cite{aurpa2022reading}. These systems are designed to interpret human language queries and provide relevant answers based on their understanding of the text. While some QA systems are open-domain, capable of addressing a wide variety of questions, others focus on more specific areas, known as closed-domain systems \cite{mervin2013overview}. In this study, we turn our attention to a closed-domain QA system designed to answer questions in Bengali, that is a language with limited NLP resources \cite{sen2022bangla}.

Bengali is spoken by over 230 million people and is the seventh most spoken language in the world \cite{das2022banglaser}. However, despite its widespread use, Bengali lacks the extensive NLP resources available to languages like English or Chinese. As a consequence, the development of tools for Bengali, especially in specialized areas such as question answering, has historically lagged behind other languages. This lack of resources has created a gap, particularly in addressing more complex tasks like closed-domain QA.

In recent years, though, advancements in NLP have introduced powerful transformer models like BERT \cite{9528178}, which have revolutionized language understanding across many languages. Yet, despite these global advancements, the application of such models to low-resource languages like Bengali has been relatively slow. BERT-Bangla \cite{sagorsarker2021banglabert}, a pre-trained model specifically designed for Bengali, offers a valuable opportunity to enhance QA systems for the language. Nevertheless, when applied to closed-domain QA, BERT-Bangla still struggles with corner cases and domain-specific challenges. This motivates our approach to leverage BERT-Bangla’s strengths while addressing its limitations in closed-domain applications.

To fill this gap, we employ and fine-tune BERT-Bangla for the specific task of answering questions about Khulna University of Engineering \& Technology (KUET). As one of the leading engineering institutions in Bangladesh, KUET has a wealth of data valuable to students, researchers, and faculty. Focusing on this domain allows us to develop a more specialized dataset and create a QA system tailored to handle queries about KUET’s academic programs, admissions, and infrastructure.

The primary objective of this research is to develop an effective closed-domain QA system that can efficiently handle KUET-related queries. Unlike open-domain systems, which require broader knowledge, closed-domain systems demand deep, domain-specific understanding and specialized datasets \cite{badugu2020study}. While these systems can be more accurate in their specific areas, they often struggle when faced with ambiguous or poorly structured questions \cite{9038332}. Our system addresses these challenges by fine-tuning BERT-Bangla on a carefully curated dataset, intending to improve accuracy and overcome the common limitations seen in previous Bengali QA systems.
The key contributions of this paper are as follows:
\begin{itemize}
    \item Development of a comprehensive dataset of 2500 question-answer pairs in Bengali related to KUET, ensuring coverage of common queries about the institution.
    \item Fine-tuning of BERT-Bangla for closed-domain question answering, demonstrating the model's ability to handle specific vocabulary and nuances of Bengali text.
    \item Performance evaluation using standard metrics like Exact Match (EM) and F1 score, with a focus on improving the model's ability to understand and process complex queries in a specialized domain.
\end{itemize}

The rest of the paper is organized as follows.
Section II presents the literature review. Section III describes the methodology. Section IV illustrates the experimental analysis. Finally, section V concludes the paper with several future directions.

\section{Literature Review}
The authors of \cite{9528178} introduced BERT-Bangla, a language model specifically pre-trained on a large corpus of unlabeled Bangla text, aimed at enhancing informative question-answering systems in Bangla. They addressed challenges encountered in developing such systems for resource-limited languages. By evaluating BERT-Bangla on various Bangla NLP classification tasks, they achieved superior performance compared to existing Bangla language models. Additionally, they developed BQuAD, a Bangla Question Answering Dataset containing question-answer pairs across multiple domains. Fine-tuning BERT-Bangla on BQuAD resulted in an effective extractive question-answering model capable of providing precise answers from user-provided Bangla contexts.

Besides this, the authors of \cite{9038332} present the Bangla Informative Question Answering System (BIQAS), a machine learning technique leveraging Bengali Natural Language Processing (BNLP) to help users find relevant information. This research employs three mathematical and statistical methods: cosine similarity, Jaccard similarity, and the Naive Bayes algorithm, focusing on question-answering data. The cosine similarity method integrates with the Singular Value Decomposition (SVD) dimension reduction technique to enhance efficiency in space and time complexity. The study is divided into two main components: data preprocessing and establishing relationships between user questions and informative questions. The results demonstrate high accuracy, achieving 93.22\% with cosine similarity, 84.64\% with Jaccard similarity, and 91.31\% with the Naive Bayes algorithm.

Alongside these efforts, the authors in \cite{aurpa2022reading} developed a question-answering system based on reading comprehension (RC) for the Bangla language, addressing a significant gap in existing NLP research, where no RC dataset had previously been available. They constructed a dataset comprising 3,636 reading comprehensions, complete with associated questions and answers. To derive accurate answers from this dataset, they employed a transformer-based deep neural network model, utilizing architectures such as LSTM, Bi-LSTM with attention, RNN, ELECTRA, and BERT. Among these, BERT and ELECTRA demonstrated superior performance. The BERT model achieved an impressive 87.78\% testing accuracy and 99\% training accuracy, while ELECTRA yielded training and testing accuracies of 82.5\% and 93\%, respectively.

In addition to this, the work \cite{tahsin2021deep} addressed the lack of a large-scale question-answering (QA) dataset and pre-trained models for Bengali, despite its global significance as the seventh most spoken language. They utilized state-of-the-art transformer models to develop a QA system based on a synthetic reading comprehension dataset, which was translated from the widely-used English benchmark, SQuAD 2.0. Additionally, they gathered a smaller, human-annotated QA dataset from Bengali Wikipedia, featuring topics relevant to Bangladeshi culture, for model evaluation. The study compared the performance of their models against human children through survey experiments to establish a baseline score for Bengali QA systems.

Furthermore, the authors of \cite{keya2020bengali} introduced a context-based question-answering (QA) system for Bengali, utilizing a deep learning-based Seq2Seq model trained on a general knowledge dataset. In this system, the context and question serve as the input to the encoder, while the corresponding answer is generated by the decoder. The model employs LSTM cells to maintain the sequence of input and output tokens. The research focused on expanding QA capabilities for Bengali, a language underrepresented in AI development compared to others. Using a dataset of 2,000 Bengali general knowledge entries, the model achieved 99\% training accuracy and 89\% validation accuracy, demonstrating strong performance in answer prediction.

Moreover, the authors of \cite{9254680} present an initial effort toward building a question-answering (QA) framework for the Bangla language, focusing on providing explicit answers rather than a list of references. Recognizing the lack of existing datasets, they created their own Bangla QA dataset. The proposed model incorporates a layered approach using several deep learning-based sub-models to understand context and generate appropriate answers for various question types. Their model, based on LSTM, achieved an F1 score of 92.16\% (considering partial matches) and an exact match score of 76.8\% on the test dataset.

Finally, the authors of \cite{9084028} proposed a closed-domain factoid question-answering system for the Bengali language, addressing the challenges it faces in computational linguistics despite being widely spoken. The system extracts answers by combining multiple sources and achieved an accuracy of 66.2\% when the object name was mentioned, and 56.8\% without it. Additionally, the system was able to identify relevant documents containing the answer with 72\% accuracy. The sub-components, such as the question and document classifier, demonstrated accuracy rates of 90.6\% and 75.3\% across five coarse-grained categories.

\section{Methodology}
\begin{figure*}
    \centering
    \includegraphics[width=.75\linewidth]{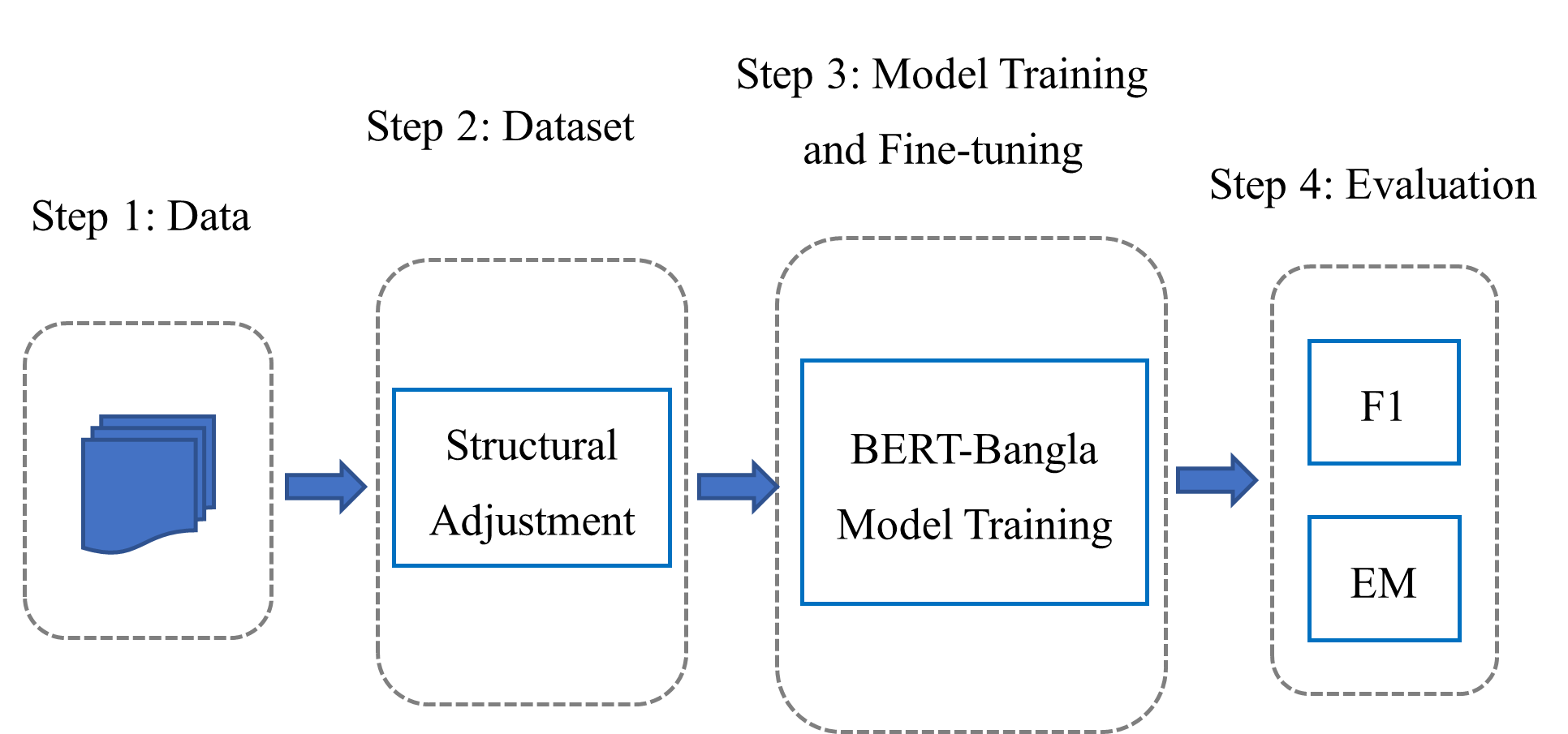}
    \caption{Workflow diagram or the framework.}
    \label{fig:workflow}
\end{figure*}

This section outlines the approach taken to develop the Bengali closed-domain question-answering system, from data collection to model fine-tuning and evaluation. The methodology is divided into the following key stages: data collection, dataset creation, fine-tuning the BERT-Bangla model, and evaluation. A workflow diagram of the framework is shown in fig. \ref{fig:workflow}.

\subsection{Data Collection}
The data collection process began with identifying sources relevant to Khulna University of Engineering \& Technology (KUET). Two main sources were selected: KUET’s official website and Wikipedia articles related to this institution. Data from KUET’s website includes information on academic programs, faculty, admissions, research facilities, and student services. Wikipedia served as a secondary source to provide general information about the institution, helping to create a more comprehensive dataset.

A web scraping script was developed to extract text data from these sources. After scraping, the data was manually reviewed to ensure relevance. The final dataset consisted of clean and relevant text passages about KUET, which were then used to generate question-answer pairs. The data extraction process focused on ensuring that the data captured different aspects of KUET’s academic and administrative functions, making it suitable for training the model on a wide range of queries.

\subsection{Dataset Creation}
Once the data was collected, we created a structured dataset following the pattern used in SQuAD v1.1 \cite{rajpurkar_squad_explorer}. The dataset consists of 100 context paragraphs, each selected from the scraped text data. For each context, 2 to 5 question-answer pairs were generated, resulting in a total of 2500 question-answer pairs. These pairs were carefully crafted to cover the common queries that prospective students or other stakeholders might have about KUET.

\begin{figure}
    \centering
    \includegraphics[width=1\linewidth]{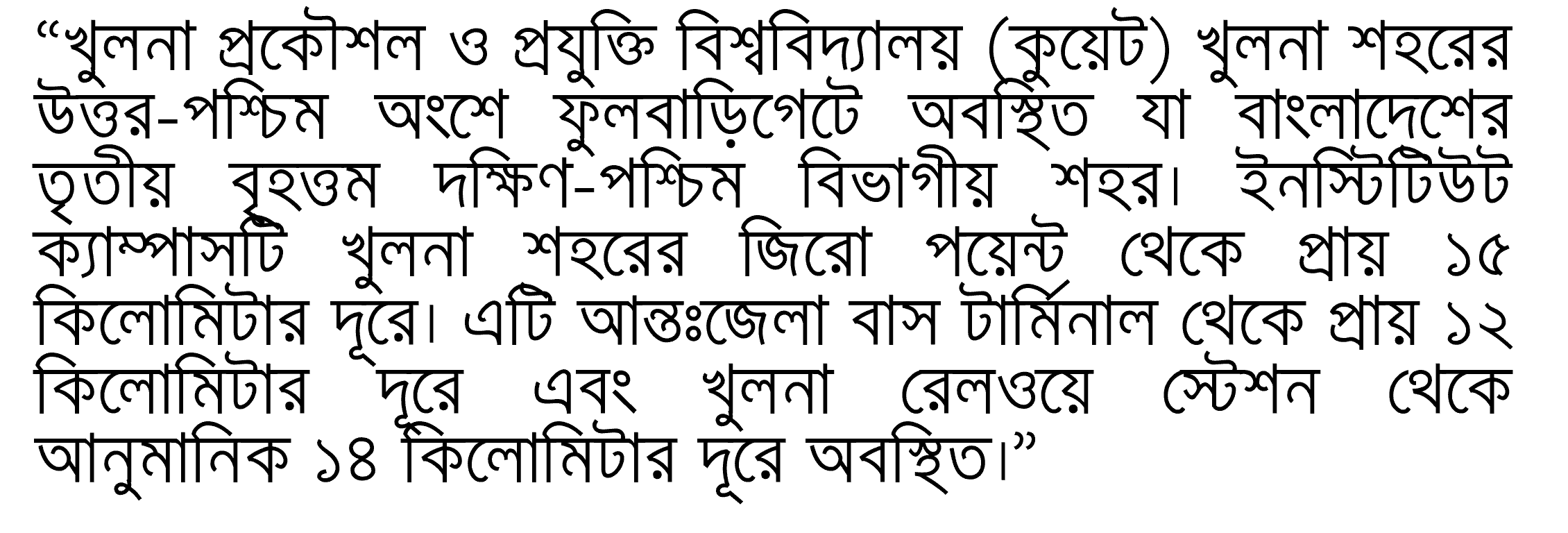}
    \caption{Sample dataset.}
    \label{fig:sampledataset}
\end{figure}

An example context from the dataset reads as of Fig. \ref{fig:sampledataset}.

\begin{figure}
    \centering
    \includegraphics[width=1\linewidth]{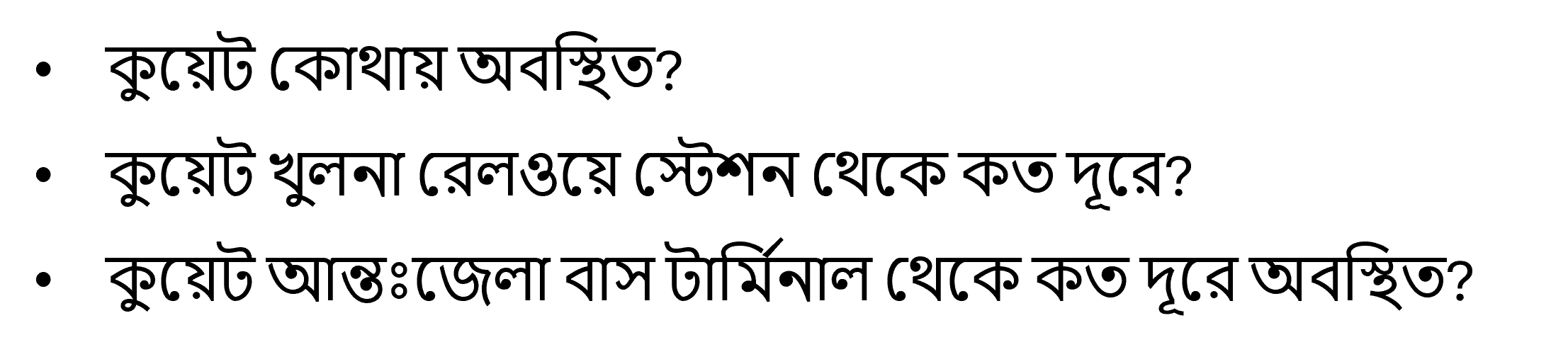}
    \caption{Sample questions.}
    \label{fig:samplequestions}
\end{figure}

The corresponding questions might include as of Fig. \ref{fig:samplequestions}.

This format ensures that the model learns to handle queries related to the location, infrastructure, and specific distances between important landmarks.

\subsection{Fine-Tuning BERT-Bangla}
\begin{figure}[t]
    \centering
    \includegraphics[width=0.5\linewidth]{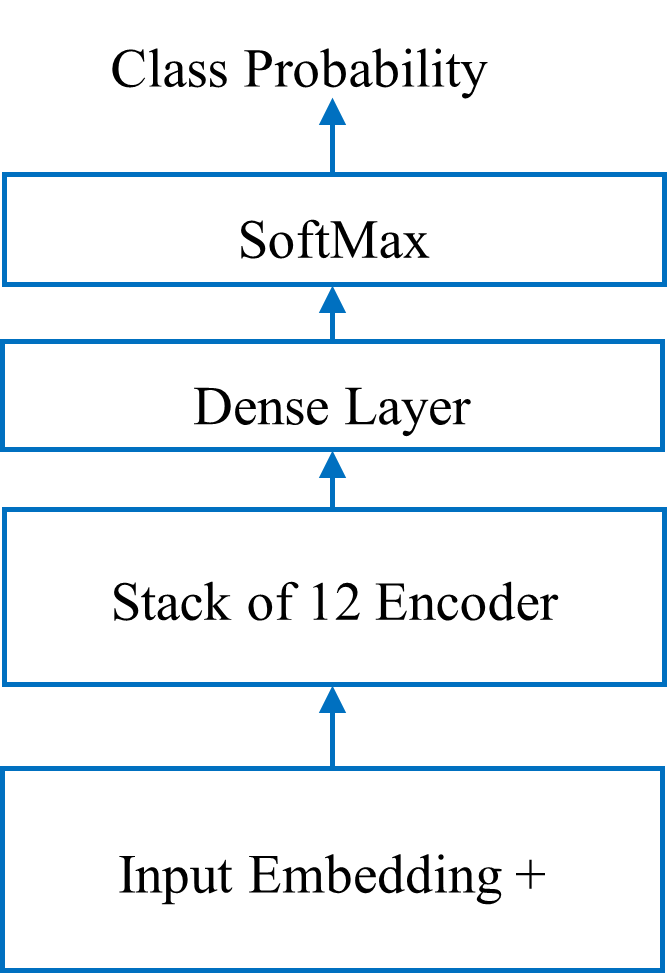}
    \caption{BERT architecture.}
    \label{fig:bertarchi}
\end{figure}

\begin{figure}
    \centering
    \includegraphics[width=.9\linewidth]{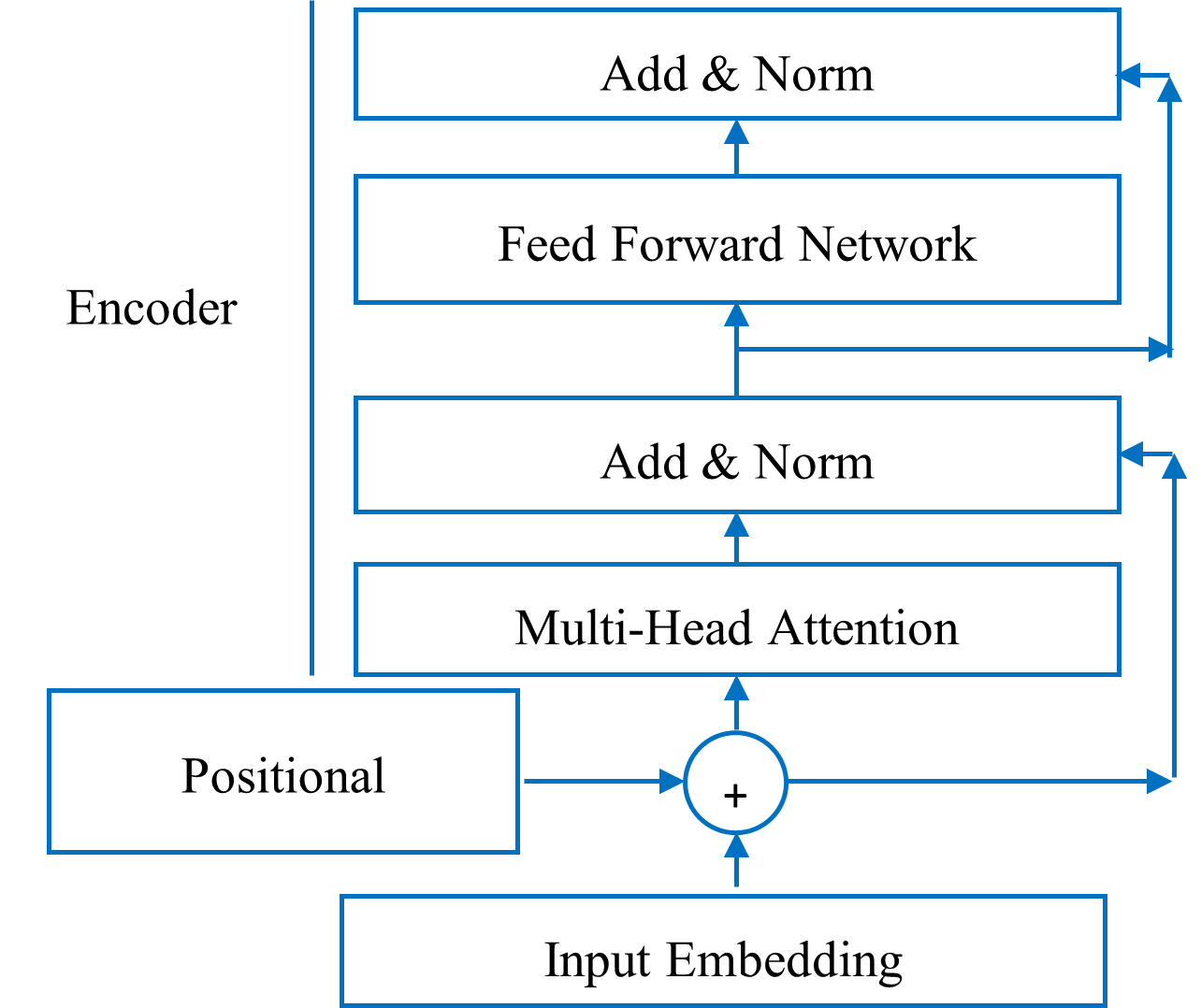}
    \caption{A single encoder layer.}
    \label{fig:encoderlayer}
\end{figure}

The next step was to fine-tune the BERT-Bangla model \cite{sagorsarker2021banglabert}, a pre-trained transformer model designed specifically for the Bengali language. BERT (Bidirectional Encoder Representations from Transformers) is a language model that uses transformers, a type of deep learning model, to understand the context and meaning of words in a sentence. BERT-Bangla was pre-trained on a large corpus of Bengali text data, making it well-suited to NLP tasks in Bengali. Fig. \ref{fig:bertarchi} shows the BERT architecture and Fig. \ref{fig:encoderlayer} shows a single Encoder layer.

\begin{figure*}[t]
    \centering
    \includegraphics[width=.75\textwidth]{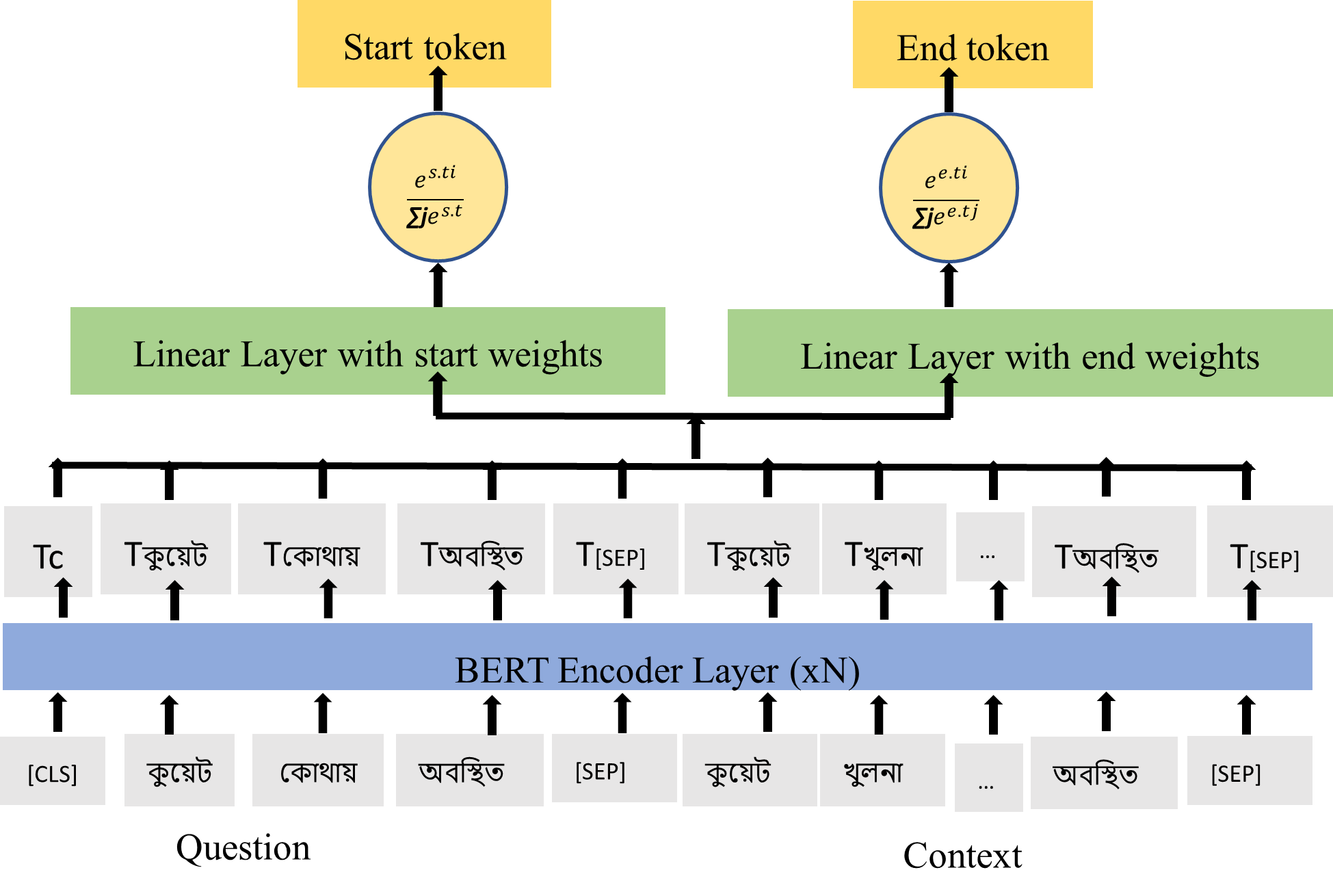}
    \caption{Fine-tuning BERT for question-answering.}
    \label{fig:Fine-tuning BERT for question-answering}
\end{figure*}

For this research, we fine-tuned BERT-Bangla on the 2500-question dataset. Fine-tuning involves adjusting the model’s weights and hyperparameters to improve its performance on a specific task—in this case, question-answering. Fig. \ref{fig:Fine-tuning BERT for question-answering} illustrates the fine-tuning of the BERT for our framework.

Besides, the following parameters were chosen during the fine-tuning process:
\begin{itemize}
    \item Learning Rate: 2e-5
    \item Batch Size: 16
    \item Training Epochs: 3
\end{itemize}
The model was trained on Google Colab using a Tesla T4 GPU, which provided sufficient computational power to handle the training process efficiently. Dropout regularization was applied during training to prevent overfitting.

\subsection{Evaluation}
After fine-tuning, the system was evaluated using three key metrics: Exact Match (EM), F1 score, and Perplexity as they are mostly used in QA systems \cite{karra2024analysis}. The EM score measures the percentage of predicted answers that exactly match the ground truth answers. F1 score balances precision and recall by considering partial overlaps between predicted and correct answers, while Perplexity measure how well a language model predicts a sequence of words. It indicates how "confused" the model is when making predictions—the lower the perplexity, the better the model is at making accurate predictions. These metrics provide insight into both the system’s accuracy and its ability to handle variations in the way questions are phrased. They can be defined as follow.

\[
EM = \frac{\text{Number of Correct Predictions}}{\text{Total Number of Questions}} \times 100
\]

\[
F1 = 2 \times \frac{\text{Precision} \times \text{Recall}}{\text{Precision} + \text{Recall}}
\]

\[
\text{Perplexity} = 2^{-\frac{1}{N} \sum_{i=1}^{N} \log_2 P(w_i)}
\]

The system was also tested for robustness by evaluating its performance on more complex queries, where the answer required understanding subtle nuances or multiple pieces of information from the context.

\section{Experimental Analysis}
This section presents the experimental setup, results of the QA system, with both quantitative and qualitative analyses of the system's performance.

\subsection{Experimental Setup}
The proposed framework was developed on a single 64-bit PC with an Intel® Core™ i5-10500 CPU running at 3.10GHz and 12 GB of RAM. Additionally, Google Colab has been used to implement the coding portion with T4 GPU for faster processing.
\subsection{Quantitative Results}
The system's performance was evaluated using the Exact Match (EM) score and F1 score. The EM score, which shows the percentage of questions answered correctly, was 55.26\%. The F1 score, which measures how well the predicted answers match the actual answers, was 74.21\%. The perplexity value was 5.71. These metrics indicate that the system can accurately answer simple questions but needs improvement for more complex ones. The results are shown in Table \ref{tab:results}.

\begin{table}[htbp]
\caption{Quantitative Analysis of the QA System}
\begin{center}
\begin{tabular}{|c|c|}
\hline
\textbf{Metric} & \textbf{Score} \\
\hline
Exact Match (EM) & 55.26\% \\
F1 Score & 74.21\% \\
Perplexity & 5.71 \\
\hline
\end{tabular}
\label{tab:results}
\end{center}
\end{table}

These results suggest that while the system performs well for simpler queries, additional improvements are needed to handle more ambiguous or context-dependent questions effectively.

\subsection{Qualitative Results}
\begin{figure}
    \centering
    \includegraphics[width=1\linewidth]{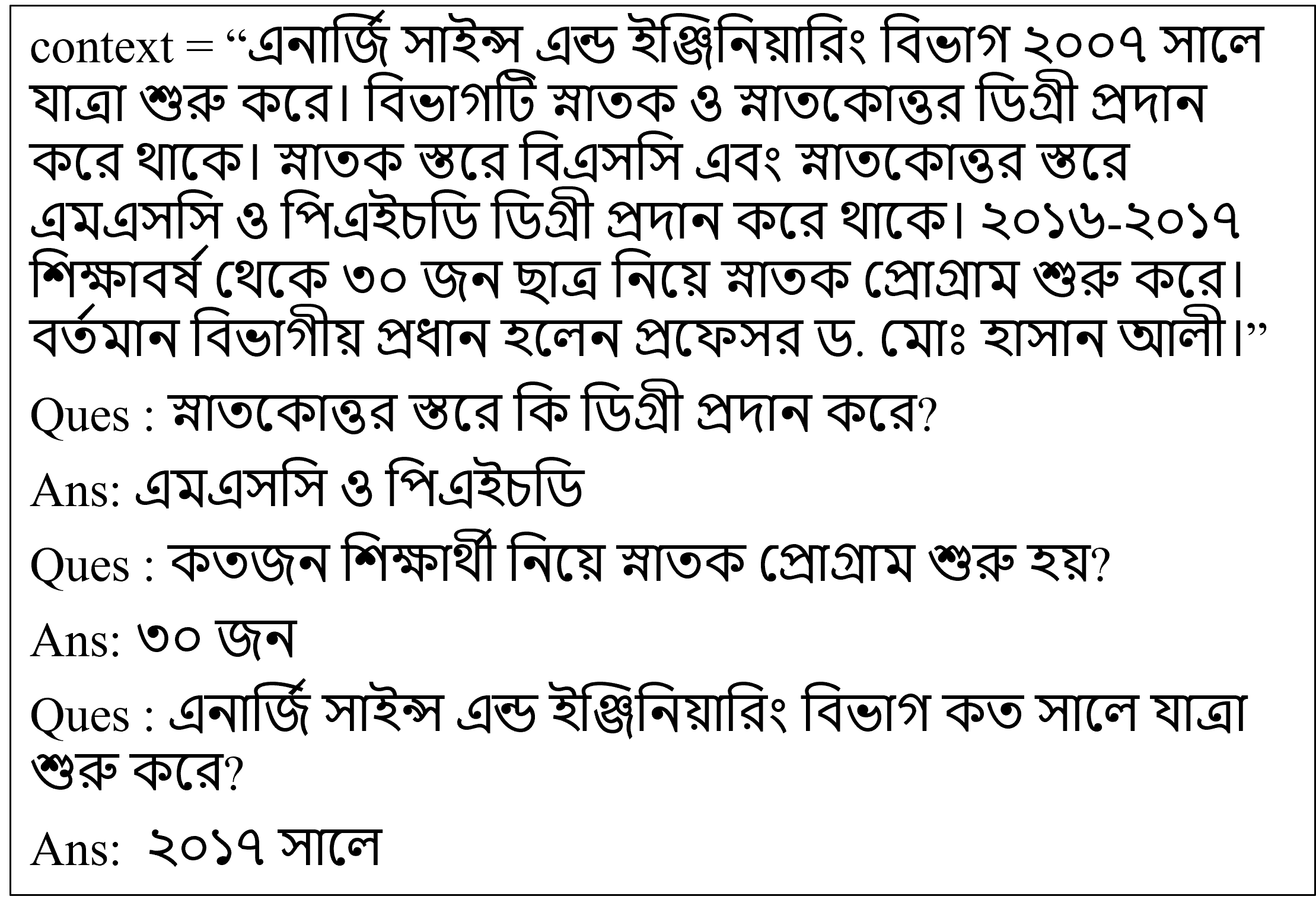}
    \caption{Sample context and query.}
    \label{fig:contextquery}
\end{figure}

To better understand the system's performance, we conducted a qualitative evaluation using several real-world examples from the KUET domain. Sample context and queries are shown in Figure. \ref{fig:contextquery}. The results demonstrate the system’s proficiency in handling factual and structured queries.

\subsection{Analysis of Results}
When examining the results, we observed that the system excels in answering questions with clear and straightforward answers derived directly from the context. The 55.26\% Exact Match (EM) score demonstrates its ability to handle fact-based queries effectively, while the 74.21\% F1 score highlights its capacity to retrieve relevant information even when an exact match isn't found. Moreover, a perplexity value of 5.71 shows that the model is reasonably adept at processing the text, although there is still potential for enhancement, particularly in handling more challenging scenarios.

That said, the system's lower performance on more complex or ambiguous questions indicates areas where further refinement is needed. Enhancing the system could involve incorporating additional data sources, optimizing the fine-tuning process, and investigating approaches for integrating external knowledge, such as knowledge graphs. By focusing on these improvements, we are confident that the system’s overall performance can be greatly enhanced.

\section{Conclusion}
In this paper, we presented a question-answering system for Bengali, fine-tuned specifically for a closed domain using BERT-Bangla. The system was evaluated on a dataset of 2500 question-answer pairs related to KUET, achieving an Exact Match score of 55.26\% and an F1 score of 74.21\%. These results demonstrate the potential of BERT-Bangla in handling domain-specific queries but also highlight areas for improvement in addressing more complex questions. Future work will focus on expanding the dataset, incorporating external knowledge, and enhancing the model’s generalization to other domains. Future work includes expanding the dataset to incorporate more diverse queries from multiple domains to improve the model's generalization. Additionally, integrating external knowledge graphs or databases will be investigated to address more complex, context-dependent queries. Another avenue of exploration is using multimodal data to enhance the system's ability to respond to queries involving both text and other data forms, such as images or audio.
\balance

\bibliographystyle{ieeetr}  
\bibliography{main}

\end{document}